\documentclass{article}
\usepackage{spconf,amsmath,graphicx}
\usepackage{subcaption}
\usepackage{blindtext}
\usepackage{multirow}
\title{Multilingual Speech Recognition using Knowledge Transfer across Learning Processes}
\name{Rimita Lahiri~$^1$$^*$\thanks{*This work was done during internship at Microsoft. The authors would like to thank Kshitiz Kumar, Jian Wu and Jinyu Li for technical discussions as well as Michael Zeng and Xuedong Huang for their project support.}, Kenichi Kumatani~$^2$, Eric Sun~$^2$, Yao Qian~$^2$}
\address{
  $^1$Signal Analysis and Interpretation Laboratory, University of Southern California, USA\\
  $^2$Microsoft Corp., USA}

\begin{document}
\ninept
\maketitle
\begin{abstract}

Multilingual end-to-end~(E2E) models have shown a great potential in the expansion of the language coverage in the realm of automatic speech recognition~(ASR). In this paper, we aim to enhance the multilingual ASR performance in two ways, 1)~studying the impact of feeding a one-hot vector identifying the language, 2)~formulating the task with a meta-learning objective combined with self-supervised learning (SSL). We associate every language with a distinct task manifold and attempt to improve the performance by transferring knowledge across learning processes itself as compared to transferring through final model parameters. We employ this strategy on a dataset comprising of 6 languages for an in-domain ASR task, by minimizing an objective related to expected gradient path length. Experimental results reveal the best pre-training strategy resulting in 3.55\% relative reduction in overall WER. A combination of LEAP and SSL yields 3.51\% relative reduction in overall WER when using language ID.

\end{abstract}
\section{Introduction}
\label{sec:intro}

In the past few decades, the speech recognition fraternity has seen remarkable technological advancement, specially in the field of multilingual ASR~\cite{toshniwal2018multilingual,NGaur2021,zhou2021configurable}. Deploying a single ASR model for multiple languages is specifically challenging because of the inherent difference between the sub-word units, lexicons and word inventories associated with different languages. These challenges led to the growing interest in learning multilingual models with shared representations across languages to circumvent burdensome explicit data requirements. 

\par Most of the early efforts~\cite{dahl2011context} in the ASR field employed 3 independent modules, namely acoustic model~(AM), language model~(LM) and pronunciation model~(PM) to solve the problem. Later, the E2E ASR models gained popularity because of their simpler structure for encapsulating AM, LM and PM all in a single network while maintaining competitive performance. In fact, most of the previous efforts in multilingual ASR have been limited to using multilingual AM employing shared hidden layers~\cite{heigold2013multilingual}, stacked bottleneck features~\cite{cui2015multilingual,sercu2017network}, multitask learning~\cite{chen2015multitask} and so on. Watanabe et al.~\cite{watanabe2017language} introduced an E2E language independent model for joint language identification and multilingual ASR tasks. Later, Kannal et al.~\cite{Kannan2019} proposed a streaming E2E multilingual model using a combination of conditioning on language vector and usage of training-language specific adapter modules.

\par Language ID carries meaningful information for language representations for downstream tasks, thus it is a key component for multilingual ASR systems. Traditional systems used to employ multiple monolingual ASR systems in parallel along with a language ID predictor module. Depending on the predicted language ID appropriate downstream transcripts used to be triggered. This approach being extremely expensive, later joint ASR-language ID modelling gained attention in the literature. Waters et al. in \cite{waters2019leveraging} use a recurrent neural network based streaming language detector and feed the detector output as auxiliary input to the encoder of a multilingual ASR system. Prior works \cite{li2019bytes, seki2018end, muller2018neural, Hou2020} have reestablished that multilingual ASR performance is usually enhanced by using auxiliary inputs for language representation. Previous promising results have prompted us to use language ID as an input for the in-domain multilingual ASR task.

\par Recently, self-supervised learning~(SSL) is shown to be very effective in the field of ASR~\cite{AConneau2006,zhang2020pushing,wang2021unispeech,zhang2021bigssl,Karimi2022}. SSL algorithms attempt at finding a good representation from unlabeled data. One of the major challenges for developing the multi-lingual ASR system is learning the cross-lingual representation for not only high-resource but also low-resource languages~\cite{AConneau2006,wang2021unispeech,Karimi2022}. SSL is particularly efficient in such cases, where a general representation is learnt by exploiting the substantial amount of unlabeled data and leveraging those learnt representations for the supervised ASR task using limited labeled data.
\par Another research direction gaining significant attention these days is meta-learning due to its success in computer vision related tasks~\cite{rusu2018meta,snell2017prototypical}. These algorithms present the learning of a new task as a learning problem itself. Usually, meta-learning problems are solved by training a meta-learner by backpropagating through the entire training process. This backpropagation through thousand of gradient steps is always prone to becoming unstable. To get rid of such issues, Flennerhag et al. introduced \textit{LEAP}~\cite{flennerhag2018transferring}, a meta-learning framework which efficiently transfers knowledge across learning processes by introducing an optimal initialization with the  shortest expected path length. The work has claimed this type of framework to yield superior performance as compared to other state of the art meta-learning methods in computer vision based tasks.
Motivated by the prior work, we combine the benefits of both the approaches in a single framework. We formulate the multilingual ASR task with a meta-learning objective of minimizing the expected path length traversed by the processes from initialization to the final set of parameters along with combining iterative self-training and pre-training in usual SSL setup. 

\par In this work, we contribute to solve the multilingual ASR task along 2 directions, firstly we investigate whether feeding language ID as the input will improve the multilingual ASR performance or not. Moreover, we focus on enhancing the ASR performance by employing a strategy combining SSL and knowledge transfer across learning processes.

The rest of the paper is organized as follows:~Section~\ref{sec:methods} describes the overall methodologies employed to boost multi-lingual ASR performance in this work. Section~\ref{sec:ex_setup} provides experimental setup, details of the dataset used in this work. Key outcomes of the experiments are tabulated and interpreted in section~\ref{results}. Finally, section 5 provides conclusions and highlights possible future extensions.

%\cite{Rebuffi17}
%\cite{Kannan2019}
%\cite{Houlsby2019}
%\cite{Hou2020}
%\cite{Hou2021}

\section{Methodology}
\label{sec:methods}

%This section describes the strategies we investigate to enhance the multilingual ASR performance.

\subsection{Language ID as an input}

One of the major challenges faced by multilingual ASR research fraternity is the varying availability of transcribed data from different languages. As a result, ASR models tend to be tailored to the resource rich over-represented languages. We investigate conditioning with language ID as a strategy to address the issue of imbalance in availability of labeled data.

\par The objective is to ensure that the ASR model will try to learn individual language traits based on the language ID instead of getting influenced by the languages having more data. Prior works have introduced the usage of language vectors in non-streaming E2E multilingual~\cite{toshniwal2018multilingual} and multidialect models~\cite{li2018multi}. The language identification information can be fed in numerous ways like simple one hot vector, language specific learnt embeddings or as clusters trained using cluster adaptive training ~(CAT)~\cite{tan2015cluster} and so on. Prior works~\cite{toshniwal2018multilingual,Kannan2019} have reported that employing simple one hot vectors yields competitive performance as compared to the more complex methods. Following that, in this work we study the impact of using language information by concatenating the language representation with the input features while feeding to the neural network architecture.

\subsection{Self-supervised learning (SSL)}

\begin{figure*}[t]
  \centerline{\includegraphics[width=0.9\textwidth, scale = 0.5]{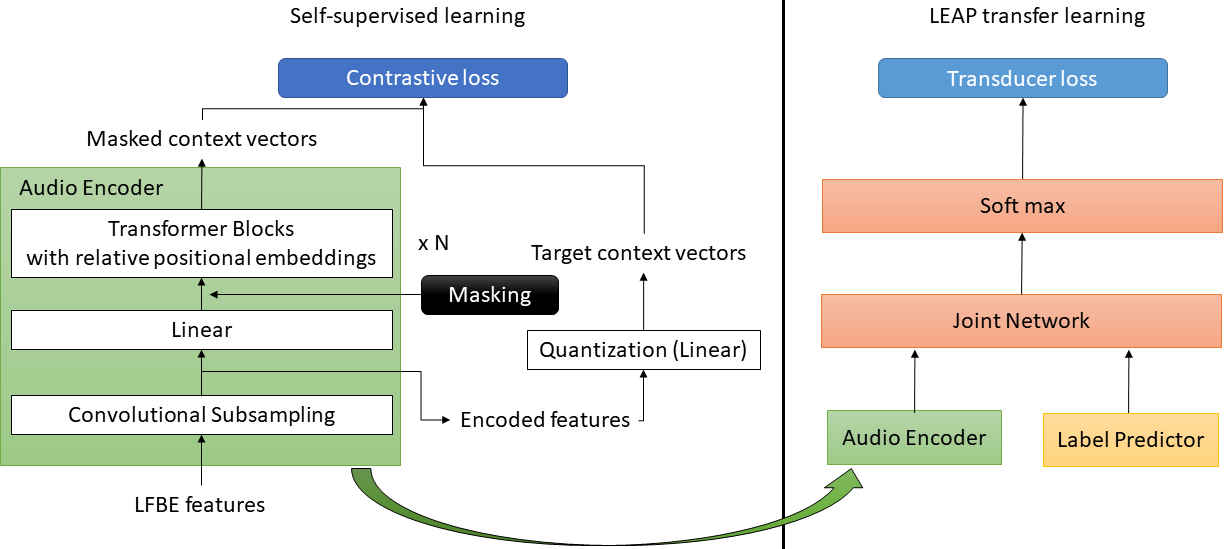}}
  \vspace{-0.3cm}
  \caption{Multi-lingual self-supervised learning (SSL)}
  \label{fig:ssl}
\end{figure*}

To obtain a good multi-lingual representation, we employ the \emph{log-filter-bank-energy-to-vector} (lfbe2vec) similar with~\cite{zhang2020pushing,Karimi2022} as the SSL step. Figure~\ref{fig:ssl} illustrates the schematic block chart of the lfbe2vec. As shown in figure~\ref{fig:ssl}, initial time steps are sampled randomly and downsampled with a convolutional network followed by the linear transformation. We mask the subsequent 10 time steps with the mask probability of 0.065. The masked features are fed to the Transformer encoder to yield masked context vectors. In this work, we use the Transformer with relative positional embeddings~\cite{li2020comparison} instead of Conformer used in ~\cite{zhang2020pushing}. Negative samples are drawn randomly from same utterance but other positions of target vectors. Finally lfb2vec optimizes the contrastive loss between masked positions of context vector and target vectors. We use multi-lingual data for SSL. The pretrained encoder is connected to the transducer decoder that consists of the label predictor, joint network, uni-directional LSTMs and Softmax layer. The whole Transformer transducer will be further optimized with the LEAP described in the next section.  

\subsection{Transferring knowledge across task manifolds}

LEAP is a meta-learning framework that accumulates all the information related to the task geometries observed during the training process. In contrast to traditional knowledge transfer learning methods that only consider final parameters, this framework aggregates the task manifolds throughout the learning process, thus avoiding any information loss. To know the worth of an initialization, the notion of path length is considered as an objective criterion. In this case, Euclidean distance is not very accurate, as this metric ignores the trajectory of the learning process. Therefore, any notion of length associated with the gradient-based trajectory on the loss surface can be able to encode all the information related to the learning task efficiently.

\par Ideally when the gradients are updated in the same direction, that indicates a smooth task unlike the harder tasks where there are frequent updates of gradients in opposite directions making it undesirable. The LEAP framework utilizes this insight, by formulating a meta-learning objective aiming to minimize the expected length of the gradient descent trajectory across tasks. The framework aims to find the initialization resulting in the shortest path for all the task while considering all the constraints related to the task geometries throughout the training process.

\par Let us first denote the input and target sample as $x$ and $y$, respectively. A task $\tau = (f_\tau, p_\tau, u_\tau)$ can be defined as a learning process to obtain the mapping of $x \rightarrow y$ based on the samples drawn from the distribution $p_\tau(x,y)$. It starts with a random initialization $\theta_\tau^0 = \theta^0$ and gradually progresses across iterations using the update rule $u_\tau$ following $\theta_\tau^{i+1} = u_\tau(\theta_\tau^i)$. Assuming that it requires $K_\tau$ gradient updates to minimize the objective function $f_\tau$, the sequence ${\{\theta_\tau^i\}}_{i=0}^{K_\tau}$ specifies the approximate trajectory traversed on the task manifold $M_\tau$ with distance $d(\theta_0;M_\tau)$. Given a distribution of tasks $p(\tau)$, each of the candidate solution will be related to a specific expected gradient path length defined as $E_{\tau \sim {p(\tau)}}[d(\theta_0;M_\tau)]$ denotes the quality of the initialization. The candidate initialization with the shortest expected gradient path length is likely to transfer maximum knowledge across learning processes and is considered to be \emph{pareto} optimal in that sense.

\par Starting from a second best candidate solution $\psi^0$, this framework first finds the baseline gradient trajectories $\{ \psi^i_\tau \}_{i=0}^{K_\tau}$, for each task $\tau$ in a batch $B$. Since all the tasks share the same initialization $\psi^0_\tau = \psi^0$. These baselines are used to update the gradient path distance metric using the equation, where $\bar \gamma_\tau^{i} = (\psi_\tau^i,f(\psi_\tau^i))$ is the frozen forward point from the baseline and $\gamma_\tau^{i} = (\theta_\tau^i,f(\theta_\tau^i))$ is the point on the updated gradient path initialiazed at $\theta^0$.

\begin{equation}
    \bar d_p(\theta^0;M_\tau,\psi_\tau) = \sum_{i=0}^{K_\tau - 1}{\|\bar \gamma_\tau^{i+1} - \gamma_\tau^i \|}^p_2
\end{equation}

Formally, the above distance metric encodes all the constraints, optimizing $\theta^0$ with respect to $\psi$ pulls the initialization forward along each task specific gradient path and the objective can be stated as,

\begin{equation}
\begin{aligned}
    \displaystyle \min_{\theta^0} \bar F(\theta^0;\psi) & =  E_{\tau \sim {p(\tau)}}[\bar d(\theta^0;M_\tau,\psi_\tau)] \\
    \textrm{s.t.} & \; \theta_\tau^{i+1} = u_\tau(\theta_\tau^i), \; \theta_\tau^0 = \theta^0 \\
\end{aligned}
\end{equation}

\par
\begin{figure}[t]
    \centering
    \includegraphics[width=0.4\textwidth, scale = 0.5 ]{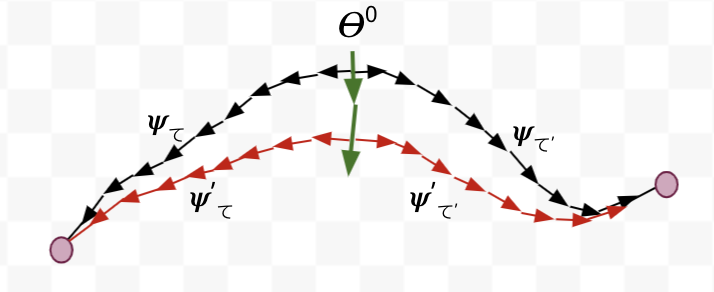}
    \vspace{-1.5ex}
    \caption{Illustration of working principle of LEAP}
    \label{fig:LEAP_img}
\vspace{-2.0ex}
\end{figure}

 To have a better understanding of how this framework operates, let us consider 2 distinct tasks $\tau$ and $\tau '$ as shown in Fig.~\ref{fig:LEAP_img}. 

\begin{figure}[h]
    \centering
    \includegraphics[width=0.48\textwidth, scale = 1.5 ]{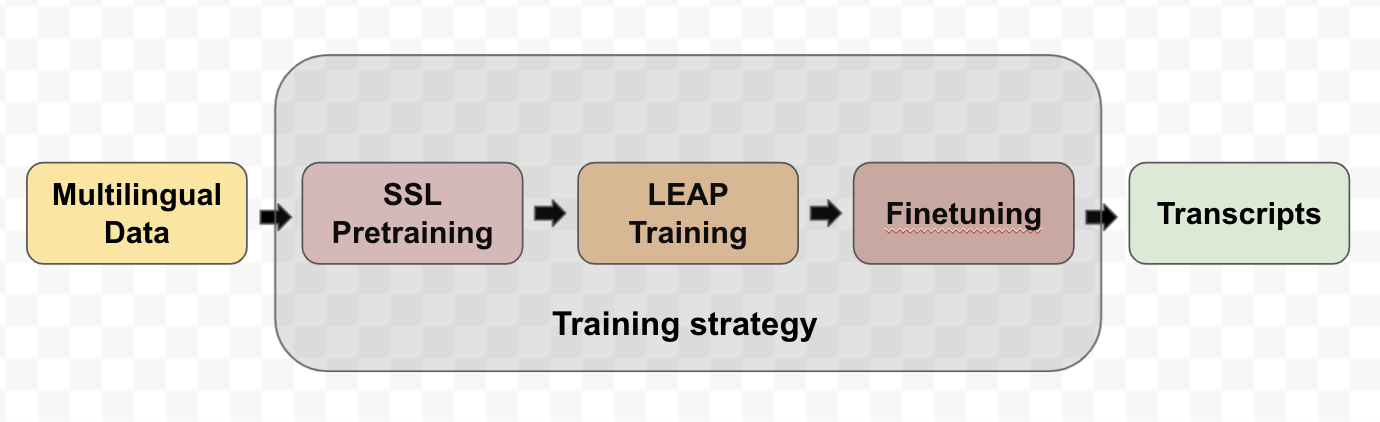}
    \vspace{-2.0ex}
    \caption{Overall training strategy}
    \label{fig:training_img}
\vspace{-2.0ex}
\end{figure}

The extent of knowledge transfer across these tasks depends on the suitability of the initialization. It starts from an initialization $\theta^0$  where the tasks $\tau$ and $\tau '$ generates the trajectory paths $\psi_\tau$ and $\psi_{{\tau} '}$ respectively. Gradually, LEAP follows the green path to improve the initialisation with an objective to minimize the expected gradient path length. 
For example, the improved initialization has expected shorter task trajectories (denoted by red arrow-lines in Fig.~\ref{fig:LEAP_img}) ${\psi '}_\tau$ and ${\psi '}_{{\tau} '}$ for tasks $\tau$ and $\tau '$ respectively. If 2 candidate initialization solutions yield equivalent performance in each task, the initialization with shortest expected gradient path trajectory encodes maximum knowledge sharing.

\subsection{Training strategy}

In this work, we consider each of the 6 languages as a distinct task. Accordingly, we expect the LEAP to provide the optimum model over 6 languages. We follow a 3 step training recipe shown in Fig. \ref{fig:training_img} to incorporate the benefits of the LEAP framework on top of SSL. In the first step, we pre-train the network with an aim to minimize contrastive loss between masked and target context vectors. Next, we fine-tune the pre-trained model using the LEAP meta-learning framework and further fine-tune the model obtained from the previous step to minimize the normal Transformer Transducer~(T-T) loss~\cite{li2020comparison,QZhang2020}. The motivation behind incorporating the LEAP framework was to find the optimal initialization point for the tasks which is likely to facilitate smoother training trajectory in the following steps.

% \subsection{Model architecture}

% The ASR model is a sequence transducer comprising an LSTM decoder and a conformer encoder. The building blocks of a conformer encoder is a stack of conformers which comprises of a deck of a multiheaded self attention, depth wise convolution and fully connected layers. We use a combination of pre-training and iterative self training~\cite{zhang2020pushing} for the baseline experiments. 

% \par A variant of wav2vec 2.0~\cite{baevski2020wav2vec} is used for the pre-training phase. Instead of raw waveforms, here logmel spectrograms are fed into the network. In this work, similar to ~\cite{baevski2020wav2vec} the pre-training is carried out with an objective to minimize the contrastive loss between the masked and target context vectors. Here, the target context vectors are obtained by passing the encoded features through a linear layer instead of a quantization module.

% \par For noisy student training, firstly the unlabeled dataset is jointly transcribed using the pre-trained ASR model and a language model. The extracted transcripts and the labeled transcripts are used in the following step to train the new pre-trained model generated using specaugment. Thus a teacher model is first used to generate pseudo labels for the unlabeled dataset, which is then used to train the next generation student models and the cycle repeats where every succeeding model is trained using the outputs generated from the previous generation model.

\section{Experimental Setup}
\label{sec:ex_setup}
\subsection{Dataset}
\begin{table}[t]
%%\begin{subtable}{0.5\linewidth}
%%\centering
%%\captionsetup{justification=centering}
%%\begin{tabular}{cc}
%%\hline
%%\hline
%% Language & Hrs \\
%%\hline
%%\hline
%% German(DE) & 1420    \\
%%\hline 
%% Italian(IT) & 2011   \\
%%\hline
%%Russian(RU) &  764  \\
%%\hline
%%Spanish(ES) &  2084  \\
%%\hline
%%French(FR) &  1733  \\
%%\hline
%%Portuguese(PT) &  2579  \\
%%\hline
%%\hline
%%\end{tabular}
%%\subcaption{Duration of subset of 6 language data with 6.5k BPEs (10,591 hours)}
%%\label{tab:data10k}
%%\end{subtable}%
%%\begin{subtable}{0.5\linewidth}
\centering
\captionsetup{justification=centering}
\begin{tabular}{cc}
\hline
\hline
 Language & Hrs \\
\hline
\hline
 German (DE) &  5688   \\
\hline 
 Italian (IT) & 8049   \\
\hline
Russian (RU) &  3055  \\
\hline
Spanish (ES) & 8332  \\
\hline
French (FR) &  6930 \\
\hline
Portuguese (PT) & 10318  \\
\hline
\hline
\end{tabular}
%%\subcaption{Duration of subset of 6 language data with 6.5k BPEs (42,372 hours)}
%%\label{tab:data42k}
%%\end{subtable}
\vspace{-0.6em}
\caption{Duration of Training data: 6 language data with 6.5k BPEs (42,372 hours)}
\label{tab:traindata}
\vspace{-2.0ex}
\end{table}

\begin{table}[t]
\centering
\captionsetup{justification=centering,margin=1cm}
\resizebox{0.48\textwidth}{!}{
\begin{tabular}{cc cc}
\hline
\hline
 Language & Code & Number of utterances & Number of words \\
\hline
\hline
German & DE & 48708	& 446215 \\
\hline
Italian & IT & 24881	& 211163 \\
\hline
Russian & RU & 6328	& 108736 \\
\hline
\multirow{2}*{French} & FR-CA & 17708 & 178392	\\
\cline{2-4}
 & FR-FR & 25371 & 273150 \\
\hline
\multirow{2}*{Spanish} & ES-MX & 19712 & 267438 \\
\cline{2-4}
 & ES-ES & 23199 & 291183  \\
\hline
\multirow{2}*{Portuguese} & PT-BR & 18643 & 262894  \\
\cline{2-4}
 & PT-PT & 2560 & 44070   \\
\hline
\hline
\end{tabular}
}
\vspace{-0.6em}
\caption{Test dataset details}
\label{tab:testdata}
\vspace{-3.0ex}
\end{table}

%...

\begin{table*}[h]
	\centering
	\begin{tabular}{|c||c|c|c|c|c|c|c|c|c||c|}
	    \hline
	    Experiments & DE & IT & RU & FR-FR & FR-CA & ES-ES & ES-MX & PT-PT & PT-BR & Overall \\ \hline\hline
	    No pretraining without Lang ID & 20.44 & 18.74 & 32 & 25.03 & 21.98 & 18.52 & 20.81 & 23.19 & 22.1 & 21.65\\ \hline  
        %Conventional SSL without Lang ID &  &  & &  & & & & & \\ \hline
        %LEAP-SSL without Lang ID &  &  & &  & & & & & \\ \hline
        No pretraining with Lang ID &  17.79 & 16.76 & 25.26 & 19.85 & 22.02 & 16.18 & 18.54 & 21.39 & 21.62 & 19.13 \\ \hline 
        %% SSL only with Lang ID & 19.16 & 19.90 & 28.17 & 23.20 & 21.08 & 17.12 & 19.58  & 22.23 & 23.28 \\ \hline
        LEAP-SSL with Lang ID & 17.12 & 16.40 & 25.22 & 19.31 & 20.87 &  15.41 & 17.55 & 21.68 & 20.83 & 18.45 \\ \hline
	\end{tabular}
	\vspace{-0.6em}
	\caption{Evaluation of multi-lingual pretraining methods in the in-domain language task.}
    \label{tab:leap_results_42k}
\end{table*}

\begin{table*}[h]
   %\begin{minipage}{\textwidth}
	\centering
	\begin{tabular}{|c||c|c|c|c|c|c|c|c|c||c|}
	    \hline
	           Experiments & DE & IT & RU & FR-FR & FR-CA & ES-ES & ES-MX & PT-PT & PT-BR & Overall \\ \hline\hline
	    No pretraining with Lang ID & 19.93 & 20.05 & 28.73 & 24.37 & 22.07 & 18.17 & 20.08 & 24.92 & 23.04 & 21.43 \\ \hline
        SSL only with Lang ID & 20.27  & 19.47 & 28.86 & 23.6 & 21.83 & 17.98 & 19.97 & 24.1 & 22.87 & 21.25\\ \hline
        LEAP-SSL with Lang ID  & 19.19 & 18.67 & 28.08 & 22.76 & 21.14 & 17.58 & 19.84 & 23.81 & 23.05 & 20.67\\ \hline
        %No pretraining without Lang ID & \textbf{56.94}  & \textbf{53.83} & \textbf{67.29} &  \textbf{60.41} & \textbf{57.63} & \textbf{57.21} & \textbf{59.38} & \textbf{73.38} & \textbf{65.11} \\ \hline 
        No pretraining without Lang ID & 25.18  & 23.88 & 35.04 & 25.27 & 29.61 & 22.60 & 24.66 & 27.54 & 26.26 & 25.71 \\ \hline 
        SSL only without Lang ID & 23.26 & 22.2 & 31.64 & 27.83 &  24.08 & 20.59 & 22.97 & 24.97 & 23.59 & 23.92\\ \hline
        LEAP-SSL without Lang ID & 23.03 & 22.38 & 32.42 & 29.19 & 24.01 & 21.04 & 23.65 & 25.52 & 24.83 & 24.42\\ \hline
	\end{tabular}
	\vspace{-0.6em}
	\caption{Effect of language ID input for SSL methods with a subset of data.}
    \label{tab:leap_results_10k}
\end{table*}
% Subset of 6 language data with 6.5k BPEs (10,591 hours) 
% DE  1,420 
% ES  2,084 
% FR 1,733 
% IT 2,011 
% PT 2,579 
% RU 764 

%Subset of 6 language data with 6.5k BPEs (42,372 hours) 
%DE  5,688 
%ES  8,332 
%FR 6,930 
%IT 8,049 
%PT 10,318 
%RU 3,055 

% lang. & #utterances & #words
% DE 48708	446215	19.16
% IT 24881	211163	19.90
% RU 6328	108736	28.17
% FR-CA 17708	178392	23.20	
% FR-FR 25371	273150	21.08
% ES-MX 19712	267438	19.58
% es-ES 23199	291183	17.12
% PT-BR 18643	262894	22.23
% PT-PT 2560	44070	23.28
% \caption{Stats of the test data.}

The dataset used for training consists of 6 languages totaling 42,372 hours. As shown in Table \ref{tab:traindata}, our training dataset is comprised of data from German, Italian, French, Spanish, Portuguese and Russian. It is worth noting here that we report the experimental results for multiple variants of Portuguese (PT-BR and PT-PT), French (FR-FR and FR-CA) and Spanish (ES-ES and ES-MX). Table~\ref{tab:testdata} tabulates the statistics of our test set for each locale: number of utterances and number of words. The dataset of each locale also contains not only various speakers but also different speech recognition tasks such as command-and-control tasks in mobile, office and car scenarios, Cortana phrases, dictation, video indexing, conversational speech in telecommunication or meetings and so on. 

For training a network, we further split the whole training data set into training and validation sets in order to determine the convergence. The amount of training data for each language is different as shown in Table \ref{tab:traindata}. We, thus, sample the lower resource data more frequently to balance the language data distribution during training.

\subsection{Implementation details}

We used a 80-dimensional LFBE feature extracted at an interval of every 10\textit{ms}. The Transformer transducer architecture consists of the Transformer encoder and LSTM decoder. The Transformer encoder comprises of 2 convolutional layers followed by 18 blocks of the relative positional Transformers. Each Transformer block contains 2~$\times$~2048 dimensional feed-forward layers, a multi-head attention with 8 heads and relative positional embeddings. Embedding dimension is fixed to be 512. The ASR decoder consists of 2 blocks of uni-directional LSTM layers followed by a 1024 dimensional feed forward layer. The Adam optimizer is used for SSL pretraining and AdamW is used for conventional fine-tuning. For LEAP-SSL fine-tuning, Adam is used as the meta optimizer to minimize the expected path length while Stochastic Gradient Descent~(SGD) is used for minimizing the T-T loss for individual tasks. In SSL, the numbers of model updates were 25,000 and 325,000 for the warm-up and linear decay stage, respectively. For fine-tuning, we performed model updates until the loss on the validation data converged. The same batch size was used for all the fine-tuning experiments. 

\section{Results and Analysis}
\label{results}

We first investigate how much the combination of SSL and LEAP pretraining can improve the  multi-lingual ASR performance. Table~\ref{tab:leap_results_42k} shows the word error rates (WERs) for each method in each row and for each locale in each column. As baselines, Table~\ref{tab:leap_results_42k} shows the recognition performance without pretraining. The overall WER in the last column of the table indicates the WER average weighted with the number of words in each language set. It is clear from Table~\ref{tab:leap_results_42k} that LEAP-SSL pretraining can provide better accuracy overall. It is also clear that the use of the language ID one-hot vector leads to better accuracy. It is interesting to note that for specific locales, using language ID as auxiliary input has shown a significant improvement~(21.06\% and 20.69\% relative reduction in WER for RU and FR-FR locales respectively) while in some cases there is no significant improvement~(PT-BR, FR-CA locales). 

Moreover, we analyze the effect of the use of language ID information in the case that pretraining is employed. Table~\ref{tab:leap_results_10k} shows the WERs for each method and locale. For quickly generating the numbers in Table~\ref{tab:leap_results_10k}, we used a quarter of the training data set for each language. It is clear from Table~\ref{tab:leap_results_10k} that the use of the language ID one-hot vector improves recognition accuracy in each method. By comparing the results without pretraining to those with SSL only, we can see that the SSL method itself can improve multi-lingual recognition accuracy. The improvement of SSL is especially prominent in the case that the input language ID is unknown. Interestingly, we use the same data for unlabeled and labeled training in these experiments. We are accordingly led to believe that SSL with multiple language data can generate the better multi-lingual representation. It is also clear from Table~\ref{tab:leap_results_10k} that LEAP combined with SSL can provide an additional improvement with the language ID input. The observations reported in Table~\ref{tab:leap_results_10k} illustrate the superior performance of LEAP-SSL with the use of language ID in all the locales except PT-BR. Contrarily, the experiments without using the language ID have shown an opposite trend in terms of WERs, for most of the locales the performance is slightly degraded with using LEAP-SSL. One possible explanation can be without language information, the LEAP framework is unable to obtain an optimal initialization, but it is worth investigating more for understanding the cause of the performance drop.

\section{Conclusion}

In this paper, we have investigated different strategies for enhancing multilingual ASR performance for an in-domain language task. In particular, we focus on feeding language ID one hot vector and minimizing the expected gradient path length across task manifolds. We report lower WERs in the experiments using language ID for all the languages considered for our in-domain multilingual ASR task. We incorporate the LEAP meta-learning framework in the SSL scheme with an added objective of minimizing the expected gradient path length across the language based tasks. Combining LEAP with SSL yields enhanced performance for all the languages in terms of the WER.

\par In the future, we plan to carry out an extensive set of experiments on the entire dataset and its subset to have a thorough understanding of the potential of using the adopted strategies. Moreover, we will try to analyze the efficiency of these training strategies by comparing with other state of the art meta-learning paradigms. Extending the current work for a larger set of languages also seems to be a fascinating research direction to be opted in future.

\bibliographystyle{IEEEbib}
\bibliography{strings}

\begin{thebibliography}{10}

\bibitem{toshniwal2018multilingual}
Shubham Toshniwal, Tara~N Sainath, Ron~J Weiss, Bo~Li, Pedro Moreno, Eugene
  Weinstein, and Kanishka Rao,
\newblock ``Multilingual speech recognition with a single end-to-end model,''
\newblock in {\em Proc. ICASSP}. IEEE, 2018, pp. 4904--4908.

\bibitem{NGaur2021}
Neeraj Gaur, Brian Farris, Parisa Haghani, Isabel Leal, Pedro~J Moreno, Manasa
  Prasad, Bhuvana Ramabhadran, and Yun Zhu,
\newblock ``Mixture of informed experts for multilingual speech recognition,''
\newblock in {\em Proc. ICASSP}, 2021.

\bibitem{zhou2021configurable}
Long Zhou, Jinyu Li, Eric Sun, and Shujie Liu,
\newblock ``A configurable multilingual model is all you need to recognize all
  languages,''
\newblock in {\em Proc. ASRU}, 2021.

\bibitem{dahl2011context}
George~E Dahl, Dong Yu, Li~Deng, and Alex Acero,
\newblock ``Context-dependent pre-trained deep neural networks for
  large-vocabulary speech recognition,''
\newblock {\em IEEE Transactions on audio, speech, and language processing},
  vol. 20, no. 1, pp. 30--42, 2011.

\bibitem{heigold2013multilingual}
Georg Heigold, Vincent Vanhoucke, Alan Senior, Patrick Nguyen, Marc’Aurelio
  Ranzato, Matthieu Devin, and Jeffrey Dean,
\newblock ``Multilingual acoustic models using distributed deep neural
  networks,''
\newblock in {\em Proc. ICASSP}. IEEE, 2013, pp. 8619--8623.

\bibitem{cui2015multilingual}
Jia Cui, Brian Kingsbury, Bhuvana Ramabhadran, Abhinav Sethy, Kartik Audhkhasi,
  Xiaodong Cui, Ellen Kislal, Lidia Mangu, Markus Nussbaum-Thom, Michael
  Picheny, et~al.,
\newblock ``Multilingual representations for low resource speech recognition
  and keyword search,''
\newblock in {\em Proc. ASRU}. IEEE, 2015, pp. 259--266.

\bibitem{sercu2017network}
Tom Sercu, George Saon, Jia Cui, Xiaodong Cui, Bhuvana Ramabhadran, Brian
  Kingsbury, and Abhinav Sethy,
\newblock ``Network architectures for multilingual speech representation
  learning,''
\newblock in {\em Proc. ICASSP}. IEEE, 2017, pp. 5295--5299.

\bibitem{chen2015multitask}
Dongpeng Chen and Brian Kan-Wing Mak,
\newblock ``Multitask learning of deep neural networks for low-resource speech
  recognition,''
\newblock {\em IEEE/ACM Transactions on Audio, Speech, and Language
  Processing}, vol. 23, no. 7, pp. 1172--1183, 2015.

\bibitem{watanabe2017language}
Shinji Watanabe, Takaaki Hori, and John~R Hershey,
\newblock ``Language independent end-to-end architecture for joint language
  identification and speech recognition,''
\newblock in {\em Proc. ASRU}. IEEE, 2017, pp. 265--271.

\bibitem{Kannan2019}
Anjuli Kannan, Arindrima Datta, Tara~N. Sainath, Eugene Weinstein, Bhuvana
  Ramabhadran, Yonghui Wu, Ankur Bapna, Zhifeng Chen, and Seungji Lee,
\newblock ``{Large-Scale Multilingual Speech Recognition with a Streaming
  End-to-End Model},''
\newblock in {\em Proc. Interspeech}, 2019.

\bibitem{waters2019leveraging}
Austin Waters, Neeraj Gaur, Parisa Haghani, Pedro Moreno, and Zhongdi Qu,
\newblock ``Leveraging language id in multilingual end-to-end speech
  recognition,''
\newblock in {\em Proc. ASRU}. IEEE, 2019, pp. 928--935.

\bibitem{li2019bytes}
Bo~Li, Yu~Zhang, Tara Sainath, Yonghui Wu, and William Chan,
\newblock ``Bytes are all you need: End-to-end multilingual speech recognition
  and synthesis with bytes,''
\newblock in {\em Proc. ICASSP}. IEEE, 2019, pp. 5621--5625.

\bibitem{seki2018end}
Hiroshi Seki, Shinji Watanabe, Takaaki Hori, Jonathan Le~Roux, and John~R
  Hershey,
\newblock ``An end-to-end language-tracking speech recognizer for
  mixed-language speech,''
\newblock in {\em Proc. ICASSP}, 2018, pp. 4919--4923.

\bibitem{muller2018neural}
Markus M{\"u}ller, Sebastian St{\"u}ker, and Alex Waibel,
\newblock ``Neural language codes for multilingual acoustic models,''
\newblock {\em arXiv preprint arXiv:1807.01956}, 2018.

\bibitem{Hou2020}
W.~Hou, Y.~Dong, B.~Zhuang, L.~Yang, J.~Shi, and T.~Shinozaki,
\newblock ``Large-scale end-to-end multilingual speech recognition and language
  identification with multi-task learning,''
\newblock in {\em Interspeech}, 2020.

\bibitem{AConneau2006}
Alexis Conneau, Alexei Baevski, Ronan Collobert, Abdelrahman Mohamed, and
  Michael Auli,
\newblock ``Unsupervised cross-lingual representation learning for speech
  recognition,''
\newblock {\em CoRR}, vol. abs/2006.13979, 2020.

\bibitem{zhang2020pushing}
Yu~Zhang, James Qin, Daniel~S Park, Wei Han, Chung-Cheng Chiu, Ruoming Pang,
  Quoc~V Le, and Yonghui Wu,
\newblock ``Pushing the limits of semi-supervised learning for automatic speech
  recognition,''
\newblock {\em arXiv preprint arXiv:2010.10504}, 2020.

\bibitem{wang2021unispeech}
Chengyi Wang, Yu~Wu, Yao Qian, Kenichi Kumatani, Shujie Liu, Furu Wei, Michael
  Zeng, and Xuedong Huang,
\newblock ``Unispeech: Unified speech representation learning with labeled and
  unlabeled data,''
\newblock {\em Proc. ICME}, 2021.

\bibitem{zhang2021bigssl}
Yu~Zhang, Daniel~S. Park, Wei Han, James Qin, Anmol Gulati, Joel Shor, Aren
  Jansen, Yuanzhong Xu, Yanping Huang, Shibo Wang, Zongwei Zhou, Bo~Li, Min Ma,
  William Chan, Jiahui Yu, Yongqiang Wang, Liangliang Cao, Khe~Chai Sim,
  Bhuvana Ramabhadran, Tara~N. Sainath, Françoise Beaufays, Zhifeng Chen,
  Quoc~V. Le, Chung-Cheng Chiu, Ruoming Pang, and Yonghui Wu,
\newblock ``Bigssl: Exploring the frontier of large-scale semi-supervised
  learning for automatic speech recognition,''
\newblock 2021.

\bibitem{Karimi2022}
Mostafa Karimi, Changliang Liu, Kenichi Kumatani, Yao Qian, Tianyu Wu, and Jian
  Wu,
\newblock ``Deploying self-supervised learning in the wild for hybrid
  automaticspeech recognition,''
\newblock in {\em Submitted to ICASSP 2022}, 2022.

\bibitem{rusu2018meta}
Andrei~A Rusu, Dushyant Rao, Jakub Sygnowski, Oriol Vinyals, Razvan Pascanu,
  Simon Osindero, and Raia Hadsell,
\newblock ``Meta-learning with latent embedding optimization,''
\newblock {\em arXiv preprint arXiv:1807.05960}, 2018.

\bibitem{snell2017prototypical}
Jake Snell, Kevin Swersky, and Richard~S Zemel,
\newblock ``Prototypical networks for few-shot learning,''
\newblock {\em arXiv preprint arXiv:1703.05175}, 2017.

\bibitem{flennerhag2018transferring}
Sebastian Flennerhag, Pablo~G Moreno, Neil~D Lawrence, and Andreas Damianou,
\newblock ``Transferring knowledge across learning processes,''
\newblock {\em ICLR}, 2019.

\bibitem{li2018multi}
Bo~Li, Tara~N Sainath, Khe~Chai Sim, Michiel Bacchiani, Eugene Weinstein,
  Patrick Nguyen, Zhifeng Chen, Yanghui Wu, and Kanishka Rao,
\newblock ``Multi-dialect speech recognition with a single sequence-to-sequence
  model,''
\newblock in {\em Proc. ICASSP}. IEEE, 2018, pp. 4749--4753.

\bibitem{tan2015cluster}
Tian Tan, Yanmin Qian, and Kai Yu,
\newblock ``Cluster adaptive training for deep neural network based acoustic
  model,''
\newblock {\em IEEE/ACM Transactions on Audio, Speech, and Language
  Processing}, vol. 24, no. 3, pp. 459--468, 2015.

\bibitem{li2020comparison}
Jinyu Li, Yu~Wu, Yashesh Gaur, Chengyi Wang, Rui Zhao, and Shujie Liu,
\newblock ``On the comparison of popular end-to-end models for large scale
  speech recognition,''
\newblock in {\em Proc. Interspeech}, 2020.

\bibitem{QZhang2020}
Qian Zhang, Han Lu, Hasim Sak, Anshuman Tripathia, Erik McDermott, Stephen Koo,
  and Shankar Kumar,
\newblock ``Transformer transducer: A streamable speech recognition model with
  transformer encoders and {RNN-T} loss,''
\newblock in {\em Proc. ICASSP}, 2020.

\end{thebibliography}

\end{document}